\documentclass[conference]{IEEEtran}
\IEEEoverridecommandlockouts

\usepackage{cite}
\usepackage{amsmath,amssymb,amsfonts}
\usepackage{algorithmic}
\usepackage{graphicx}
\usepackage{multirow}
\usepackage{textcomp}
\usepackage{comment}
\usepackage{tabularx}
\usepackage{xcolor}
\usepackage{colortbl}
\usepackage{soul}
\usepackage[T1]{fontenc}
\usepackage[utf8]{inputenc}

\def\BibTeX{{\rm B\kern-.05em{\sc i\kern-.025em b}\kern-.08em
    T\kern-.1667em\lower.7ex\hbox{E}\kern-.125emX}}

\begin{document}

\title{The Cognitive Cost of LLM Writing: A Reading-Time Comparison with Human Text}


\author{
    \IEEEauthorblockN{
        Kamer Cekini\IEEEauthorrefmark{1}\IEEEauthorrefmark{2}, Elisabetta Biondi\IEEEauthorrefmark{1}, Chiara Boldrini\IEEEauthorrefmark{1}, Andrea Passarella\IEEEauthorrefmark{1}, Marco Conti\IEEEauthorrefmark{1}
    }
    \IEEEauthorblockA{\IEEEauthorrefmark{1} IIT-CNR, via Moruzzi 1, Pisa, Italy}
    \IEEEauthorblockA{\IEEEauthorrefmark{2} University of Pisa, Pisa, Italy}
    \IEEEauthorblockN{\textit{\{name.surname\}@iit.cnr.it}}
}

\maketitle

\begin{abstract}
Although linguistic differences between human-written and Large Language Model (LLM)-generated texts have been extensively studied, their implications for the cognitive effort involved in reading remain less explored. In psycholinguistics, reading time is commonly used as an indirect measure of cognitive processing effort. Building on this perspective, we investigate whether human-written texts and their LLM-generated rewrites exhibit differences in predicted reading time that may reflect differences in processing demands for readers. To this end, we introduce RoBERTa-RT, a model trained on human eye-tracking data to predict word-level reading times. We evaluate the model on both internal and external test sets, construct a paired dataset of 496 human-written articles and their LLM-generated rewrites, and apply RoBERTa-RT to compare their predicted reading times. Our findings show that LLM-generated rewrites tend to be shorter and have slightly lower predicted total reading times, while exhibiting higher predicted reading times per word. These differences are associated with systematic linguistic changes, including longer and less frequent words, greater lexical diversity, and higher contextual predictability. From a cognitive perspective, our findings suggest that LLM-generated writing may redistribute reading demands: reducing the amount of text to process while increasing the predicted processing effort associated with individual words. This highlights the importance of considering not only the linguistic characteristics of machine-generated content, but also its potential implications for human cognitive processing.
\end{abstract}


\section{Introduction}\label{introduction}

In recent years, the use of Large Language Models (LLMs) for text generation has become increasingly widespread, driven by their ability to produce written content quickly and with limited user input \cite{wu2025survey,noy2023productivity}. Advances in their generative capabilities have also substantially improved the linguistic quality and sophistication of the resulting texts \cite{herbold2023largescale}. These developments have prompted growing research interest in the linguistic characteristics that distinguish human-written texts from those generated by LLMs. Previous studies have identified systematic differences in lexical choices, syntactic structures, grammatical patterns, and linguistic predictability \cite{munozortiz2024contrasting,reinhart2025llms,liao2023medical}. These differences have provided valuable insights into the distinctive characteristics of human-written and LLM-generated texts and have also been explored as potential signals for automatic detection \cite{wu2025survey,przystalski2026stylometry}.

Beyond these linguistic differences, an important question is whether human-written and LLM-generated texts also differ in the cognitive effort they require from readers. Reading difficulty depends not only on the information conveyed by a text but also on how that information is expressed \cite{rayner1998eye,just1980theory}. Linguistic properties such as word frequency, word length, and contextual predictability are known to influence the ease with which words are processed during reading \cite{rayner1986lexical,juhasz2003intercorrelated,rayner2004frequency,smith2013predictability}. Differences in these properties may therefore affect how readers process human-written and LLM-generated texts, even when the texts convey approximately the same information.

Reading time has long been used in psycholinguistic research to investigate the cognitive effort involved in language processing \cite{just1980theory,rayner1998eye}. Although reading time provides only an indirect measure of cognitive effort, longer reading times are generally associated with greater processing difficulty. Building on this perspective, we investigate whether human-written texts and their LLM-generated rewrites differ in the reading difficulty they may present, despite conveying approximately the same information. We also examine which linguistic characteristics may contribute to these differences. Importantly, the cognitive implications of LLM-generated writing may involve competing effects. More concise texts reduce the amount of material readers must process, potentially lowering the overall time required for reading. However, changes in lexical choices and linguistic structure may increase the processing demands associated with individual words. Understanding this balance is important for assessing whether the efficiency gains offered by LLM-generated writing extend beyond text production to the experience of the reader. 

In this work, we address two research questions:
\begin{enumerate}
    \item Do human-written texts and their LLM-generated rewrites differ in reading difficulty?
    \item Which linguistic differences between human-written and LLM-generated texts are associated with these variations in reading difficulty?
\end{enumerate}
To address these questions, we adopt a two-stage approach. In the first stage, we develop a model for estimating reading times from text. Directly measuring reading times for a large collection of texts would require extensive eye-tracking experiments involving multiple participants, making such analyses difficult to scale. We therefore develop RoBERTa-RT, a model trained on existing human eye-tracking data to predict reading times at the level of individual word occurrences. We evaluate its predictive performance on both internal and external test data before applying it to previously unseen texts. This approach enables a large-scale comparison of predicted reading times without requiring additional eye-tracking experiments for each collection of documents.

In the second stage, we construct a paired dataset of 496 Human--LLM text pairs and use RoBERTa-RT to compare the predicted reading times of human-written articles and their corresponding LLM-generated rewrites. For each human-written article, the LLM is instructed to preserve the informational content of the original article while remaining free to modify its linguistic formulation, including lexical choices, sentence structure, and text organization. This paired design allows us to compare alternative linguistic realizations of approximately the same information while reducing differences attributable to topic and content.  For each pair, we compare predicted reading times and examine several linguistic properties relevant to reading difficulty, including word length, word frequency, contextual predictability, and lexical diversity. We then investigate how differences in these properties are associated with variations in predicted reading time, seeking to identify linguistic characteristics that may help explain the observed patterns.

This work makes three main contributions. First, we extend Human–LLM comparisons to predicted reading times, providing a reader-centered complement to traditional linguistic analyses. Second, we develop and validate RoBERTa-RT, a model trained on human eye-tracking data to predict word-level reading times. Third, we construct a paired Human–LLM dataset in which each generated text is a rewrite of a specific human-written source, enabling comparisons between different linguistic realizations of approximately the same informational content.

Our results show that the differences between human-written and LLM-generated texts cannot be reduced to the simple claim that one is uniformly easier to read than the other. LLM-generated rewrites tend to be shorter and have slightly lower predicted total reading times, yet they also exhibit higher predicted reading times per word. This pattern is accompanied by systematic linguistic differences: LLM-generated texts tend to use longer and less frequent words and display greater lexical diversity, while being more predictable in context. Taken together, these findings point to a trade-off between conciseness and predicted per-word reading cost: LLM-generated rewrites use fewer words while exhibiting, on average, longer predicted reading times per word.


The remainder of the paper is organized as follows. Section~\ref{background-and-related-work} reviews related work on reading-time modeling and comparisons between human-written and LLM-generated texts. Section~\ref{sec:reading-time-prediction-model} describes the development and validation of RoBERTa-RT. Section~\ref{humanllm-comparative-analysis} introduces the paired Human--LLM dataset, the reading-time and linguistic measures, and the statistical methods employed in the analysis. Section~\ref{results} presents the results, focusing on differences in predicted reading times, linguistic characteristics, and their relationships. Finally, Section~\ref{conclusion} concludes the paper.

\section{Background and Related Work}\label{background-and-related-work}

Psycholinguistic research has long relied on reading times as an indirect measure of the cognitive effort involved in language comprehension. Longer reading times are generally associated with greater processing difficulty, making them a useful indicator of the demands imposed by a text on the reader \cite{just1980theory,rayner1998eye}. Research has identified several linguistic factors that systematically influence reading behavior. At the lexical level, word length and frequency play an important role, with longer and less frequent words typically requiring more processing time \cite{juhasz2003intercorrelated,rayner1986lexical}. Beyond individual word properties, reading times are also sensitive to contextual predictability: words that are less expected given the preceding context tend to be processed more slowly \cite{rayner2004frequency}. This relationship is commonly examined through surprisal, a measure of how unexpected a word is in its context, which has been shown to correlate systematically with reading times \cite{smith2013predictability}.

These findings have also motivated computational approaches aimed at predicting reading-time measures directly from textual input. By learning the relationship between linguistic properties and observed reading behavior, such models make it possible to estimate reading times without collecting new eye-tracking data. Singh et al. \cite{singh2016quantifying}, for example, developed feature-based supervised models trained on human eye-tracking data to predict word-level reading measures, demonstrating how these predictions could be used to assess sentence complexity and text readability. Extending this line of research to longer texts, Weller et al. \cite{weller-etal-2020-dont} investigated the prediction of document-level reading times, identifying text length as a particularly strong predictor.

Subsequent research has explored a wider range of modeling strategies, including approaches based on pretrained language models. The CMCL 2021 Shared Task on Eye-Tracking Prediction \cite{hollenstein2021cmcl} provided a common benchmark for this research, bringing together 13 teams to predict five token-level eye-tracking measures. Participating systems included traditional machine-learning models using linguistic and psychometric features, Transformer-based neural architectures, and hybrid approaches. Among these, Bestgen \cite{bestgen2021last} achieved the best overall performance using gradient-boosted decision trees with lexical and psychometric features, highlighting the continued effectiveness of feature-based methods. In contrast, Li and Rudzicz \cite{li-rudzicz-2021-torontocl} explored a neural approach based on RoBERTa, combining a regression layer with multi-stage fine-tuning. Together, these studies demonstrate that reading-time measures can be estimated from text using different computational strategies, providing a scalable alternative when direct measurements of human reading behavior are unavailable or impractical to collect.

The possibility of estimating reading times computationally is particularly relevant in the context of LLM-generated text. A growing body of research has shown that human-written and LLM-generated texts differ systematically in their linguistic properties, including lexical, grammatical, and stylistic characteristics \cite{munozortiz2024contrasting,reinhart2025llms}. Some of these differences concern properties known to influence reading behavior. For instance, Herbold et al.~\cite{herbold2023largescale} report greater lexical diversity in texts generated by GPT-4, while Liao et al.~\cite{liao2023medical} find that ChatGPT-generated texts exhibit greater predictability than human-written texts in their comparison. Such linguistic differences have also been investigated as potential indicators for automatically distinguishing machine-generated from human-written text \cite{wu2025survey,przystalski2026stylometry}. However, while these studies provide evidence of systematic differences in how human and LLM-generated texts are written, they do not establish whether these differences translate into variations in the cognitive effort required to read them.

To the best of our knowledge, differences in reading times between human-written texts and their LLM-generated rewrites have not yet been systematically investigated. To address this gap, we use computational reading-time predictions to compare paired texts with approximately equivalent informational content, examining whether predicted reading times differ and how these differences relate to linguistic characteristics. In doing so, we move beyond the analysis of linguistic differences to explore their potential implications for reading effort.

\section{Reading-Time Prediction Model: RoBERTa-RT}\label{sec:reading-time-prediction-model}

A central methodological requirement of this work is the ability to estimate reading times for texts for which no direct eye-tracking observations are available. We address this need by developing RoBERTa-RT, a RoBERTa-based model trained to predict word-level reading times directly from textual input. This section describes the eye-tracking data used for model development, the definition of the prediction target, the training and model-selection procedure, and the evaluation of the resulting model on both internal and external test data.

\subsection{Eye-Tracking Datasets}\label{eye-tracking-datasets}

The development of RoBERTa-RT relies on existing eye-tracking corpora collected by recording participants' gaze positions while they read text presented on a screen. These recordings make it possible to identify fixations, defined as intervals during which the gaze remains relatively stable on a particular region of the text, and to map them onto individual words. By summing the durations of all fixations associated with the same word occurrence, we obtain an estimate of the total time spent reading that word.


Our primary dataset is OneStop Eye Movements \cite{berzak2025onestop}, which contains eye-tracking data from 360 native English speakers reading 30 newspaper articles from \emph{The Guardian}, divided into 162 paragraphs. Texts are presented one paragraph at a time, and each paragraph is followed by a comprehension question.

We partition the 30 articles into 18 training articles, 6 validation articles, and 6 test articles. The training split is used to fit the models, the validation split to tune hyperparameters and select model configurations, and the test split is reserved for evaluating the selected model on unseen articles from the same corpus.

To increase both the amount and the diversity of data available for training, we supplement the OneStop training split with the GECO and MECO corpora. GECO \cite{cop2017geco} contains eye-tracking data collected during the reading of Agatha Christie's novel \emph{The Mysterious Affair at Styles} and includes 14 monolingual English readers and 19 Dutch--English bilingual readers. In our study, we use only the data from monolingual English participants in order to reduce variability potentially associated with differences in language proficiency. MECO \cite{siegelman2025meco}, by contrast, contains eye-tracking data collected while participants read 12 short encyclopedic texts in multiple languages, each accompanied by comprehension questions. Here too, we retain only participants whose first language is English.

Finally, we use Provo \cite{luke2018provo} exclusively as an external test set. The corpus contains 55 short English passages, averaging approximately 50 words each, read by 84 native English speakers. Because it comes from an independent study involving different participants and reading materials, it allows us to assess how well the model generalizes beyond the corpora used during development.

Table~\ref{tab:eye_tracking_datasets} summarizes the eye-tracking datasets used for model development and evaluation, reporting the number of word occurrences assigned to each split. Overall, the training set comprises 78,881 word occurrences from OneStop, GECO, and MECO, while 7,975 OneStop occurrences are reserved for validation. Final evaluation is conducted on 6,834 OneStop occurrences for internal testing and 2,743 Provo occurrences for external testing, yielding a total of 96,433 word occurrences across all splits.

\begin{table}[!t]
\centering
\caption{Eye-tracking datasets used for model training,
validation, and evaluation. Counts refer to word occurrences.}
\label{tab:eye_tracking_datasets}
\footnotesize
\setlength{\tabcolsep}{4pt}
\renewcommand{\arraystretch}{1.12}
\begin{tabular*}{\columnwidth}{@{\extracolsep{\fill}}lcr@{}}
\hline
\textbf{Dataset} & \textbf{Split} & \textbf{Word occurrences} \\
\hline
OneStop & Training & 20,364 \\
GECO    & Training & 56,410 \\
MECO    & Training & 2,107 \\
\multicolumn{2}{l}{\textbf{Total training}} & \textbf{78,881} \\
\hline
OneStop & Validation & 7,975 \\
\multicolumn{2}{l}{\textbf{Total validation}} & \textbf{7,975} \\
\hline
OneStop & Test & 6,834 \\
Provo   & External test & 2,743 \\
\multicolumn{2}{l}{\textbf{Total test}} & \textbf{9,577} \\
\hline
\multicolumn{2}{l}{\textbf{Grand total}} & \textbf{96,433} \\
\hline
\end{tabular*}
\end{table}


\subsection{Prediction Target}\label{prediction-target}

From the eye-tracking data described in the previous section, we obtain, for each participant, a reading-time measurement associated with every word occurrence in the text.

Let \(i\) denote a text and \(j\) the \(j\)-th word occurrence within that text. We denote by \(t_{ijp}\) the reading time recorded for participant \(p\) on that specific occurrence. Because each text is read by multiple participants, several reading-time measurements are available for the same word occurrence. We define the prediction target as the mean reading time across participants:

\[r_{ij}=\frac{1}{P_i}\sum_{p=1}^{P_i} t_{ijp},\]
where \(P_i\) is the number of participants who read text \(i\). Thus, \(r_{ij}\) represents the average observed reading time for the \(j\)-th word occurrence in text \(i\).

Importantly, the target is defined at the level of word occurrences, rather than word types. The same lexical item appearing in different positions or contexts is therefore treated as a distinct observation each time. This preserves information about the specific linguistic context in which the word occurs.

Given a text, the model is trained to estimate \(r_{ij}\) for each of its word occurrences. We denote the corresponding prediction by \(\hat{r}_{ij}\). For each text, we therefore obtain two aligned sequences: one containing the observed reading times, \(r_{ij}\), and one containing the predicted reading times, \(\hat{r}_{ij}\), with one value for each word occurrence.

\subsection{Models and Training}\label{models-and-training}

Having defined the prediction target, we next consider alternative modeling strategies for estimating reading times at the level of individual word occurrences. Specifically, we compare two main families of approaches: feature-based models and Transformer-based models. The feature-based models include Ridge Regression, Random Forest, CatBoost, and a Multi-Layer Perceptron (MLP). For the Transformer-based approach, we evaluate several pretrained architectures, including RoBERTa-base, RoBERTa-large \cite{liu2019roberta}, DeBERTa-v3-base, and DeBERTa-v3-large \cite{he2023debertav3improvingdebertausing}. Our goal is to compare different modeling strategies and identify the most effective approach for reading-time prediction.

For feature-based models, each word occurrence is represented by a set of features describing both the word itself and the context in which it appears. These include, for example, word length and frequency, properties of the preceding and following words, and the position of the word within the paragraph. The complete set of 15 features used by all feature-based models is reported in Table~\ref{tab:linguistic_features}. The resulting feature vector is then provided to the regression model, which learns to estimate the reading time associated with each word occurrence.

\begin{table*}[!t]
\centering
\caption{Linguistic features used for word-level reading-time prediction in feature-based models.}
\label{tab:linguistic_features}
\scriptsize
\setlength{\tabcolsep}{3pt}
\renewcommand{\arraystretch}{1.08}

\begin{tabularx}{\textwidth}{@{}lX@{\hspace{8pt}}lX@{}}
\hline
\textbf{Feature} & \textbf{Description} &
\textbf{Feature} & \textbf{Description} \\
\hline

\texttt{word\_len\_clean} &
Number of characters in the word. &
\texttt{prev\_word\_len\_clean} &
Length of the preceding word. \\

\texttt{contains\_digit} &
Presence of numerical digits (binary). &
\texttt{next\_word\_len\_clean} &
Length of the following word. \\

\texttt{has\_hyphen} &
Presence of a hyphen (binary). &
\texttt{prev\_word\_freq\_zipf} &
Zipf frequency of the preceding word. \\

\texttt{ends\_with\_punctuation} &
Ends with punctuation (binary). &
\texttt{next\_word\_freq\_zipf} &
Zipf frequency of the following word. \\

\texttt{word\_index\_in\_context} &
Word index within the paragraph. &
\texttt{prev\_ends\_with\_punctuation} &
Preceding word ends with punctuation (binary). \\

\texttt{relative\_position\_in\_context} &
Word position relative to paragraph length. &
\texttt{next\_ends\_with\_punctuation} &
Following word ends with punctuation (binary). \\

\texttt{word\_freq\_zipf} &
English word frequency on the Zipf scale. &
\texttt{gpt2\_surprisal\_bits} &
GPT-2 surprisal given preceding context (bits). \\

\texttt{is\_stopword} &
Whether the word is an English stopword (binary). &
& \\

\hline
\end{tabularx}

\vspace{2pt}
\parbox{\textwidth}{\scriptsize
\emph{Note.} All features are computed at the word level.
Word lengths exclude surrounding punctuation, and binary
features take values 0 or 1. Lexical frequencies are obtained
using \texttt{wordfreq}, stopwords using spaCy, and surprisal
using GPT-2.
}
\end{table*}


Transformer-based models, in contrast, operate directly on the text. During training, validation, testing, and subsequent inference, texts are processed one paragraph at a time. Each paragraph is tokenized and processed independently by the Transformer, which produces a contextualized representation for each subword token. Because the prediction target is defined at the word level, a single word may correspond to multiple subword tokens. In such cases, the representations of all subwords belonging to the same word are aggregated through mean pooling, yielding a single contextualized representation for each word occurrence. This representation is then passed to an MLP-based regression head, which predicts the corresponding reading time, denoted by \(\hat{r}_{ij}\). During fine-tuning, the parameters of both the Transformer encoder and the regression head are optimized jointly on the eye-tracking training data.

For all models, reading-time prediction is formulated as a regression task at the level of individual word occurrences. Let \(r_{ij}\) denote the observed mean reading time for the \(j\)-th word occurrence in text \(i\), as defined in Section~\ref{prediction-target}, and let \(\hat{r}_{ij}\) denote the corresponding model prediction. For the neural models, parameters are optimized by minimizing the Mean Squared Error (MSE) between observed and predicted reading times.

All candidate models are trained on the combined training data from OneStop, GECO, and MECO described in Section~\ref{eye-tracking-datasets}.

\subsection{Model Selection and Evaluation}\label{model-evaluation}

Although the models generate a prediction for each individual word occurrence, alternative configurations are compared at the text level. This choice reflects the downstream use of predicted reading times in our Human--LLM analyses, where word-level predictions are aggregated into text-level reading-time measures, as detailed in Section~\ref{sec:reading-time-and-linguistic-measures}. Thus, models are trained at the word level but selected according to their performance at the text level.

Specifically, for a text \(i\) containing \(n_i\) word occurrences, we define the observed mean reading time per word as
\[
y_i = \frac{1}{n_i}\sum_{j=1}^{n_i} r_{ij},
\]
and the corresponding predicted value as
\[
\hat{y}_i = \frac{1}{n_i}\sum_{j=1}^{n_i} \hat{r}_{ij}.
\]
Here, \(y_i\) represents the observed mean reading time per word for text \(i\), whereas \(\hat{y}_i\) denotes the corresponding model estimate. Given a set of \(N\) texts, performance is measured using the Mean Absolute Error (MAE):

\[
\mathrm{MAE} =
\frac{1}{N}
\sum_{i=1}^{N}
\left|
\hat{y}_i-y_i
\right|.
\]

The resulting value is expressed in milliseconds per word (ms/word) and is used as the common criterion for comparing the model configurations considered during development.


For each candidate model, we conduct hyperparameter optimization using 50 Optuna trials with the Tree-structured Parzen Estimator (TPE) sampler, aiming to minimize mean absolute error (MAE) on the OneStop validation set. The feature-based models are trained using the same set of 15 linguistic features, summarized in Table~\ref{tab:linguistic_features}. For Transformer-based models, we consider four pretrained architectures (RoBERTa-base, RoBERTa-large, DeBERTa-v3-base, and DeBERTa-v3-large), each optimized independently. 
Table~\ref{tab:model_hyperparameters} summarizes the hyperparameter optimization procedure for the evaluated models, reporting search ranges and the values selected based on performance on the OneStop validation set. 

\begin{table}[!t]
\centering
\caption{Hyperparameter search ranges and selected values
for the evaluated models.}
\label{tab:model_hyperparameters}
\scriptsize
\setlength{\tabcolsep}{2.1pt}
\renewcommand{\arraystretch}{1.10}

\begin{tabular}{@{}llcc@{}}
\hline
\textbf{Model} & \textbf{Hyperparameter} &
\textbf{Search range} & \textbf{Selected} \\
\hline

\rowcolor{gray!12}
 & Learning rate & $[2,8]\times10^{-5}$
 & $\boldsymbol{2.05\times10^{-5}}$ \\
\rowcolor{gray!12}
 & Weight decay & $[0.02,0.12]$
 & $\boldsymbol{0.0493}$ \\
\rowcolor{gray!12}
\multirow{-3}{*}{\textbf{RoBERTa-base}}
 & Warmup ratio & $[0.10,0.30]$
 & $\boldsymbol{0.2284}$ \\
\hline

\multirow{3}{*}{RoBERTa-large}
 & Learning rate & $[1,6]\times10^{-5}$ & $5.66\times10^{-5}$ \\
 & Weight decay & $[0.04,0.12]$ & $0.0405$ \\
 & Warmup ratio & $[0.15,0.35]$ & $0.3439$ \\
\hline

\multirow{3}{*}{DeBERTa-v3-base}
 & Learning rate & $[1,7]\times10^{-5}$ & $3.28\times10^{-5}$ \\
 & Weight decay & $[0.03,0.12]$ & $0.0387$ \\
 & Warmup ratio & $[0.15,0.35]$ & $0.2404$ \\
\hline

\multirow{3}{*}{DeBERTa-v3-large}
 & Learning rate & $[5,50]\times10^{-6}$ & $9.01\times10^{-6}$ \\
 & Weight decay & $[0.04,0.12]$ & $0.0772$ \\
 & Warmup ratio & $[0.15,0.35]$ & $0.2351$ \\
\hline

Ridge
 & Regularization ($\alpha$) & $[0.01,10^4]$ & $0.01$ \\
\hline

\multirow{3}{*}{Random Forest}
 & Number of trees & $[300,2000]$ & $1906$ \\
 & Min. leaf samples & $[1,30]$ & $1$ \\
 & Max. features & $[0.3,1.0]$ & $0.3246$ \\
\hline

\multirow{3}{*}{CatBoost}
 & Depth & $[3,10]$ & $10$ \\
 & Learning rate & $[0.01,0.15]$ & $0.0156$ \\
 & L2 regularization & $[1,50]$ & $10.1435$ \\
\hline

\multirow{3}{*}{MLP}
 & Hidden layers & $[1,3]$ & $2$ \\
 & Dropout & $[0,0.4]$ & $0.1000$ \\
 & Learning rate & $[10^{-5},3\times10^{-3}]$
 & $1.13\times10^{-4}$ \\
\hline

\end{tabular}
\end{table}

Using the configurations selected through this optimization procedure, we compare the candidate models on the OneStop validation set.
Table~\ref{tab:rt_model_comparison} reports the performance of the best configuration of each model, selected according to its validation MAE. Model selection is based exclusively on OneStop validation performance; OneStop Test and Provo Test are used only for post-selection evaluation.
\begin{table}[t]
\centering
\caption{Text-level MAE (ms/word) for the best configuration of each model, selected on the OneStop validation set.}
\label{tab:rt_model_comparison}
\footnotesize
\setlength{\tabcolsep}{3.8pt}
\begin{tabular}{lc|cc}
\hline
\textbf{Model} &
\textbf{OneStop Val} &
\textbf{OneStop Test} &
\textbf{Provo Test} \\
\hline
\textbf{RoBERTa-base} & \textbf{15.76} & \textbf{15.00} & \textbf{16.09} \\
RoBERTa-large         & 17.41 & 15.65 & 17.00 \\
DeBERTa-v3-base       & 17.55 & 18.29 & 20.36 \\
DeBERTa-v3-large      & 18.50 & 17.61 & 17.85 \\
\hline
MLP              & 23.77 & 18.72 & 16.74 \\
Ridge Regression & 24.63 & 19.29 & 21.10 \\
Random Forest    & 23.60 & 18.96 & 16.47 \\
CatBoost         & 27.85 & 21.49 & 22.07 \\
\hline
\end{tabular}
\end{table}
%
%
%
RoBERTa-base achieves the lowest MAE on the OneStop validation set (15.76 ms/word), outperforming all other candidate models. We therefore select the fine-tuned RoBERTa-base model for the subsequent reading-time analyses and refer to it as RoBERTa-RT.
%
Using the selected hyperparameters, RoBERTa-RT is then trained with validation-based checkpoint selection, and the checkpoint achieving the lowest OneStop validation MAE (epoch 8) is retained for final evaluation.

The selected RoBERTa-RT checkpoint was then evaluated on the two held-out test sets without further modification. On OneStop Test, the observed mean reading time is \(192.84\) ms per word, while RoBERTa-RT achieves an MAE of \(15.00\) ms/word. On Provo, which serves as an external corpus, the observed mean reading time is \(200.16\) ms per word, and the model achieves an MAE of \(16.09\) ms/word.

In both cases, the prediction error is relatively small compared with the observed mean reading times. Specifically, the MAE corresponds to approximately \(7.8\%\) of the observed mean reading time on OneStop Test and \(8.0\%\) on Provo. The comparable performance on Provo, an independently collected eye-tracking corpus, provides evidence that RoBERTa-RT generalizes beyond the data used for model development. While this evaluation does not directly establish the accuracy of small reading-time differences between individual texts, it supports the use of the model to investigate systematic patterns in predicted reading times across a large collection of Human–LLM pairs. Accordingly, the following analyses focus on aggregate differences and associations rather than precise estimates of cognitive effort for individual readers.



\section{Human--LLM Comparative Analysis}\label{humanllm-comparative-analysis}

We now use RoBERTa-RT to investigate whether human-written and LLM-generated texts differ in their predicted reading effort and linguistic characteristics. To make this comparison as controlled as possible, we first construct a paired Human--LLM corpus in which each generated text is a rewrite of a specific human-written article. We then use these pairs to compare predicted reading times and a set of linguistic properties potentially related to reading behavior.

\subsection{Human--LLM Paired Dataset}\label{humanllm-paired-dataset}

Our main research questions concern the comparison between human-written and LLM-generated texts while keeping their informational content as similar as possible. To enable such a controlled comparison, we construct a paired dataset in which each human-written article is matched with an LLM-generated rewrite of the same article. This paired design allows us to compare reading-time estimates and linguistic properties within each Human--LLM pair, reducing the influence of differences in topic or content across documents.

\paragraph{Human Source Corpus}\label{human-source-corpus}

As the source corpus, we use CCAT50 \cite{liu2006reuter}, which contains 5,000 news articles written by 50 authors, with 100 documents per author. We select 10 articles from each author, resulting in a balanced sample of 500 human-written articles. These texts serve as the source material for generating the corresponding LLM rewrites.

\paragraph{LLM Rewriting Procedure}\label{llm-rewriting-procedure}

For each of the 500 selected articles, we generate a corresponding rewrite using GPT-5.6 Terra, version dated July 9, 2026, with \emph{reasoning effort} set to \emph{medium}. Each human-written article is provided to the model in full. The following prompt is used for all articles:

\begin{quote}
``Rewrite the following English news article as a new, standalone news article in your own words. Cover the same facts and attributed statements, without adding outside information or omitting substantive content. Write naturally, using your own wording, sentence structure, and paragraphing. Return only and exclusively the rewritten article, with no introductory or explanatory text.''
\end{quote}

The prompt is designed to constrain the generated article at the informational level while leaving the model free to modify its linguistic realization. In particular, lexical choices, sentence structure, and paragraph organization may differ from the source article, while its substantive content is intended to remain comparable. This design allows us to compare alternative linguistic realizations of approximately the same informational content.

The generated texts are used exactly as returned by the model, without any manual revision or post-processing. Generation is completed successfully for 496 of the 500 source articles. In the remaining four cases, the model refuses the rewriting request because of content it identifies as sensitive; these articles are therefore excluded from subsequent analyses. The final dataset consequently contains 496 Human--LLM pairs, each consisting of one human-written article and its corresponding LLM-generated rewrite.

\subsection{Reading-Time and Linguistic Measures}\label{sec:reading-time-and-linguistic-measures}

The paired corpus allows us to investigate whether human-written and LLM-generated texts differ in the processing effort they may require from readers. At the same time, we can examine how the two types of text differ in their linguistic characteristics and whether these differences help explain the observed variation in predicted reading effort.

We begin with the reading-time perspective. We apply the RoBERTa-RT model to each of the 496 human-written articles and their corresponding LLM-generated rewrites. To accommodate RoBERTa's 512-token input limit, each article is processed paragraph by paragraph, with each paragraph fitting within the model's maximum sequence length. The resulting predictions are then combined into a sequence of predicted reading times \(\hat{r}_{ij}\) for each article, with one prediction for each word occurrence. From these word-level predictions, we derive two text-level measures: predicted total reading time and predicted mean reading time per word.

For a text \(i\) containing \(n_i\) word occurrences, the predicted \textit{total reading time} is computed as the sum of the predicted reading times across all words:

\[\sum_{j=1}^{n_i}\hat{r}_{ij}.\]
This measure represents the total time that, according to the model, would be required to read the entire article.

The predicted \textit{mean reading time per word} is instead computed by averaging the same word-level predictions:

\[\frac{1}{n_i}\sum_{j=1}^{n_i}\hat{r}_{ij}.\]
This measure captures the average predicted processing time associated with an individual word occurrence. Considering both measures is important because they capture different aspects of reading effort. Total reading time reflects the time required to communicate the entire message to a reader and is therefore strongly affected by text length. Mean reading time per word, instead, captures the average processing time associated with each word, allowing us to compare texts independently of their overall length. This distinction is particularly relevant in our setting, since the human-written articles and their LLM-generated rewrites may differ substantially in length.

We then turn to the linguistic dimension and examine a set of textual properties that may help characterize the differences between human-written and LLM-generated rewrites and interpret the observed variation in predicted reading times.  \textit{Mean word length} is computed as the average number of characters per word in each article, obtained by dividing the total number of characters across all words by the total number of word occurrences. This measure is relevant because longer words tend, on average, to require more processing time \cite{juhasz2003intercorrelated}.

\textit{Mean word frequency} is estimated using the \texttt{word\_frequency} function from the wordfreq library \cite{speer2022wordfreq}. For each word occurrence, the function returns an estimated relative frequency in English on a scale from 0 to 1. For example, a value of 0.001 corresponds approximately to one occurrence per 1,000 words. We then average these values across all word occurrences in the article. Higher values therefore indicate texts composed, on average, of more frequent words, whereas lower values indicate greater use of less common vocabulary. Lexical frequency is also closely related to reading behavior, with less frequent words generally requiring longer processing times \cite{rayner1986lexical}.

We additionally consider \textit{surprisal}, which quantifies how predictable a given word occurrence is from its preceding context. For a token \(w_t\), surprisal is defined as

\[S(w_t)=-\log P(w_t\mid w_{<t}),\]
where \(w_{<t}\) denotes the sequence of tokens preceding \(w_t\). Lower surprisal values correspond to words that are more predictable in context, whereas higher values indicate less expected words.\\
%
We estimate these probabilities using OPT-125M \cite{zhang2022opt}, a language model distinct from the one used to generate the rewrites. For each article, we compute surprisal for each evaluated token and then average these values across the text. Higher mean surprisal therefore indicates that, on average, the tokens in a text are less predictable given their preceding context. This measure is relevant because surprisal is closely related to reading time: less predictable words generally require more time to process \cite{smith2013predictability}.

Finally, we include two additional characteristics defined directly at the text level. \textit{Word count} is the total number of words in an article and serves as a measure of text length, while \textit{lexical diversity} is quantified using MTLD (Measure of Textual Lexical Diversity) \cite{mccarthy2010mtld}. Higher MTLD values indicate greater lexical diversity, whereas lower values reflect greater repetition of the same lexical items. Compared with simpler measures such as the type--token ratio \cite{mccarthy2010mtld}, MTLD was developed to be less sensitive to text length, making it more suitable for comparisons across articles of different lengths. We refer to McCarthy and Jarvis \cite{mccarthy2010mtld} for details of its computation.

\subsection{Statistical Analysis}\label{statistical-analysis}

All analyses are conducted on the 496 Human--LLM pairs described in the previous sections. Because each generated text is a rewrite of a specific human-written article, all comparisons are performed at the pair level, with each LLM-generated text compared exclusively with its corresponding human-written source.

To assess whether a given metric \(X\) differs systematically between human-written and LLM-generated texts, we compute the paired difference for each pair \(i\):

\[\Delta_i(X)=X_{\mathrm{LLM},i}-X_{\mathrm{Human},i}.\]
A positive value of \(\Delta_i(X)\) indicates that the metric is higher in the LLM-generated text, whereas a negative value indicates that it is higher in the corresponding human-written text. To assess whether human-written and LLM-generated texts differ systematically with respect to each metric, we apply a two-sided Wilcoxon signed-rank test to the 496 paired differences \(\Delta_i(X)\). The null hypothesis \(H_0\) is that there is no systematic shift between the paired observations.\\
%
%
The analysis is conducted on the seven metrics presented in Section~\ref{sec:reading-time-and-linguistic-measures}: predicted total reading time, predicted mean reading time per word, word count, mean word length, mean word frequency, mean surprisal, and MTLD. Because seven separate tests are performed, the resulting \(p\)-values are adjusted using the Bonferroni correction.

To additionally quantify the magnitude of the difference relative to the value observed in the human-written text, we compute the percentage relative change for each pair:

\[RC_i(X)=100\cdot\frac{X_{\mathrm{LLM},i}-X_{\mathrm{Human},i}}{X_{\mathrm{Human},i}}.\]
Positive values indicate an increase in the metric for the LLM-generated text relative to its human-written counterpart, whereas negative values indicate a decrease. The overall magnitude and direction of the change are summarized by the mean relative change across the 496 pairs:

\[\overline{RC}(X)=\frac{1}{N}\sum_{i=1}^{N}RC_i(X), \qquad N=496.\]

The pair-level relative changes also allow us to investigate whether changes in different metrics are associated with one another. We use Spearman's rank correlation coefficient \(\rho\), computed over the corresponding relative-change values \(RC_i(X)\) and \(RC_i(Y)\) across the 496 pairs. A positive correlation indicates that changes in the two metrics tend to move in the same direction, whereas a negative correlation indicates that they tend to move in opposite directions. We report nine Spearman correlation tests in total; their \(p\)-values are adjusted jointly using the Bonferroni correction to account for multiple comparisons. Throughout the remainder of the paper, we denote Bonferroni-adjusted \(p\)-values as \(p_{\mathrm{adj}}\).

\section{Results}\label{results}

In this section, we compare human-written texts with their corresponding LLM-generated rewrites using the methodology described in the previous section. We first examine differences in predicted reading time and text length (Section~\ref{sec:reading-time-and-text-length}). We then analyze the linguistic properties of the paired texts and investigate how changes in these properties are associated with changes in predicted reading time (Section~\ref{sec:linguistic-differences-between-human-and-llm-texts}). Finally, we examine the relationship between text compression, predicted reading time, and linguistic characteristics (Section~\ref{the-compressionreading-cost-trade-off}).

All seven Human--LLM differences examined below remain statistically significant after correction for multiple comparisons ($p_{\mathrm{adj}} < .01$). Likewise, all reported Spearman correlations remain significant after correction ($p_{\mathrm{adj}} < .01$). To avoid repetition, we therefore focus on the magnitude and direction of the observed differences and associations.

\subsection{Reading Time and Text Length}
\label{sec:reading-time-and-text-length}

As described in Section~\ref{sec:reading-time-and-linguistic-measures}, our RoBERTa-RT model provides two complementary measures of predicted reading time. Predicted total reading time estimates the time required to read and process an entire text, whereas predicted mean reading time per word captures the average processing time associated with each word.

These two measures show opposite patterns across the paired texts. LLM-generated texts have a slightly lower predicted total reading time than their human-written counterparts, averaging 115.6 s compared with 117.1 s, with a mean relative change of $\overline{RC} = -1.19\%$. By contrast, predicted mean reading time per word is higher for LLM-generated texts, averaging 233.4 ms compared with 229.9 ms for human-written texts ($\overline{RC} = +1.57\%$).

To better understand this contrast, we consider text length. LLM-generated rewrites contain, on average, 496.4 words, compared with 511.2 words in their human-written counterparts ($\overline{RC} = -2.66\%$). The shorter length of the LLM-generated rewrites offsets the higher predicted reading time per word, resulting in a slightly lower total predicted reading time for LLM-generated texts. Table~\ref{tab:reading_time_length} summarizes these differences and reports the corresponding Bonferroni-adjusted $p$-values.

\begin{table*}[t]
\centering
\small
\caption{Reading-Time and Text-Length Differences Between Human-Written and LLM-Generated Texts}
\label{tab:reading_time_length}

\begin{tabular*}{\textwidth}{@{\extracolsep{\fill}}lrrrr@{}}
\hline
\textbf{Measure} &
\textbf{Human mean $\pm$ SD} &
\textbf{LLM mean $\pm$ SD} &
\textbf{$\overline{RC} \pm$ SD (\%)} &
\textbf{$p_{\mathrm{adj}}$} \\
\hline

Total predicted RT (s)
& $117.10 \pm 25.89$
& $115.58 \pm 25.48$
& $-1.19 \pm 4.06$
& $1.45 \times 10^{-9}$ \\

Mean predicted RT/word (ms)
& $229.91 \pm 15.27$
& $233.39 \pm 14.77$
& $+1.57 \pm 2.91$
& $2.33 \times 10^{-25}$ \\

Word count
& $511.19 \pm 115.34$
& $496.39 \pm 109.40$
& $-2.66 \pm 4.49$
& $7.76 \times 10^{-34}$ \\

\hline
\end{tabular*}
\end{table*}

These results suggest that LLM-generated writing may affect reading demands in two opposing ways. By reducing text length, LLM-generated rewrites decrease the amount of material that readers must process. At the same time, their higher predicted reading time per word suggests greater processing demands at the lexical level. The resulting reduction in total predicted reading time is therefore relatively modest, despite the shorter texts. This distinction highlights why textual conciseness should not automatically be equated with cognitive ease: reducing the number of words does not necessarily reduce the processing effort associated with each word.

\subsection{Linguistic Differences  and Predicted Reading Time}
\label{sec:linguistic-differences-between-human-and-llm-texts}

To better characterize the differences between human-written and LLM-generated texts, we examine the linguistic measures introduced in Section~\ref{sec:reading-time-and-linguistic-measures}. We first compare their values across Human--LLM pairs and then investigate how changes in these measures are associated with changes in predicted mean reading time per word.

As shown in Table~\ref{tab:linguistic_differences}, LLM-generated texts differ systematically from their human-written counterparts across all four linguistic measures. Mean word length increases from 4.95 to 5.25 characters ($\overline{RC} = +6.06\%$), while mean word frequency decreases by 7.61\%, indicating a greater use of less frequent words. Lexical diversity, measured using MTLD, also increases substantially, with a mean relative change of $+31.83\%$. In contrast, mean surprisal decreases by 6.51\%, indicating that words in LLM-generated texts are, on average, more predictable from their preceding context.

\begin{table*}[t]
\centering
\small
\caption{Linguistic Differences Between Human-Written and LLM-Generated Texts}
\label{tab:linguistic_differences}

\begin{tabular*}{\textwidth}{@{\extracolsep{\fill}}lrrrr@{}}
\hline
\textbf{Measure} &
\textbf{Human mean $\pm$ SD} &
\textbf{LLM mean $\pm$ SD} &
\textbf{$\overline{RC} \pm$ SD (\%)} &
\textbf{$p_{\mathrm{adj}}$} \\
\hline

Mean word length
& $4.949 \pm 0.176$
& $5.248 \pm 0.195$
& $+6.06 \pm 2.33$
& $4.08 \times 10^{-82}$ \\

Mean word frequency
& $0.007262 \pm 0.000874$
& $0.006686 \pm 0.000790$
& $-7.61 \pm 7.34$
& $1.48 \times 10^{-61}$ \\

MTLD
& $103.02 \pm 21.11$
& $134.67 \pm 29.89$
& $+31.83 \pm 20.37$
& $1.50 \times 10^{-78}$ \\

Mean surprisal
& $2.952 \pm 0.228$
& $2.755 \pm 0.188$
& $-6.51 \pm 4.36$
& $4.80 \times 10^{-77}$ \\

\hline
\end{tabular*}
\end{table*}

To investigate how these linguistic differences relate to predicted reading time, we examine Spearman correlations between the relative changes in each linguistic measure and the corresponding changes in predicted mean reading time per word (Table~\ref{tab:linguistic_rt_relationship}).

Three characteristics show associations consistent with the higher predicted per-word reading time of LLM-generated texts. Changes in mean word length are positively correlated with changes in predicted reading time per word ($\rho = .588$), as are changes in lexical diversity ($\rho = .292$). Mean word frequency, by contrast, shows a negative correlation ($\rho = -.405$): larger decreases in lexical frequency tend to accompany larger increases in predicted reading time per word. These results indicate that increases in word length and lexical diversity, together with decreases in word frequency, are associated with higher predicted reading times per word.

\begin{table}[t]
\centering
\small
\caption{Relationship Between Linguistic Changes and Predicted Reading Time}
\label{tab:linguistic_rt_relationship}

\begin{tabular*}{\columnwidth}{@{\extracolsep{\fill}}lcc@{}}
\hline
\textbf{Linguistic measure} &
\textbf{$\rho$ with $RC_i(\mathrm{RT/word})$} &
\textbf{$p_{\mathrm{adj}}$} \\
\hline

Mean word length
& $+.588$
& $1.47 \times 10^{-46}$ \\

Mean word frequency
& $-.405$
& $4.61 \times 10^{-20}$ \\

MTLD
& $+.292$
& $2.97 \times 10^{-10}$ \\

Mean surprisal
& $+.438$
& $1.18 \times 10^{-23}$ \\

\hline
\end{tabular*}
\end{table}

A different pattern emerges for surprisal. Unlike the other linguistic characteristics, its reduction in LLM-generated texts is consistent with lower processing difficulty, as words become more predictable from their preceding context. Consistent with this interpretation, changes in surprisal are positively correlated with changes in predicted mean reading time per word ($\rho = .438$). In other words, larger reductions in surprisal tend to be associated with smaller increases in predicted per-word reading time.

The lower surprisal of LLM-generated texts may appear to contrast with their lower mean word frequency, but the two measures capture different properties. Word frequency reflects how common a word is in general language use, whereas surprisal depends on the context in which a word occurs. A text can therefore use relatively uncommon words while placing them in contexts that make their occurrence highly predictable.

Overall, the results reveal two opposing patterns. LLM-generated texts use longer and less frequent words and exhibit greater lexical diversity, with changes in these characteristics associated with higher predicted reading time per word. At the same time, their lower surprisal indicates greater contextual predictability, which is associated with smaller increases in per-word reading time. Despite this opposing pattern, LLM-generated texts exhibit a higher predicted mean reading time per word overall.

\subsection{The Compression--Reading Cost Trade-off}
\label{the-compressionreading-cost-trade-off}

While the previous section examined how linguistic changes relate to predicted reading time, we now shift the focus to text compression. Specifically, we investigate whether the extent to which LLM-generated rewrites shorten their human-written counterparts is associated with changes in per-word reading time and in their linguistic characteristics. To this end, we use the relative change in word count as a measure of compression, with more negative values indicating greater text reduction.

As shown in Table~\ref{tab:compression_correlations}, the relative change in word count is negatively correlated with the change in predicted mean reading time per word ($\rho = -.458$). Thus, pairs in which the LLM shortens the original text more substantially tend to exhibit larger increases in predicted per-word reading time.

\begin{table}[t]
\centering
\small
\caption{Spearman correlations between relative changes in word count and corresponding changes in predicted reading time and linguistic characteristics across Human--LLM pairs ($N=496$).}
\label{tab:compression_correlations}

\begin{tabular*}{\columnwidth}{@{\extracolsep{\fill}}lcc@{}}
\hline
\textbf{Measure} &
\textbf{$\rho$} &
\textbf{$p_{\mathrm{adj}}$} \\
\hline

Mean predicted RT/word
& $-.458$ & $3.46 \times 10^{-26}$ \\

Mean word length
& $-.303$ & $4.93 \times 10^{-11}$ \\

Mean word frequency
& $+.364$ & $4.47 \times 10^{-16}$ \\

MTLD
& $-.269$ & $1.10 \times 10^{-8}$ \\

Mean surprisal
& $-.362$ & $6.93 \times 10^{-16}$ \\

\hline
\end{tabular*}
\end{table}

The correlations with linguistic characteristics reveal a consistent pattern. Greater reductions in word count are associated with larger increases in mean word length ($\rho = -.303$) and lexical diversity ($\rho = -.269$), as well as larger decreases in mean word frequency ($\rho = .364$). Thus, more extensive compression tends to coincide with a stronger shift toward longer, less frequent, and more diverse vocabulary.


We also observe a negative correlation between changes in word count and surprisal ($\rho = -.362$). This indicates that greater text compression tends to be associated with smaller reductions in surprisal. Thus, although LLM-generated rewrites generally exhibit lower surprisal than their human-written counterparts, this difference becomes less pronounced as the extent of compression increases.

Taken together, these findings suggest a compression--reading cost trade-off: greater reductions in text length tend to coincide with larger increases in predicted reading time per word. This relationship is accompanied by a distinct linguistic shift toward longer and less frequent words, greater lexical diversity, and a smaller reduction in surprisal. 

\section{Conclusion}\label{conclusion}

Comparisons between human-written and LLM-generated texts have traditionally focused on linguistic differences and their potential use in identifying machine-generated content. However, these differences may also have implications for how readers process information. In this work, we explored this perspective by investigating whether human-written texts and their LLM-generated rewrites, despite conveying comparable information, differ in the reading effort they require and how these differences relate to their linguistic characteristics.

To address this question, we developed RoBERTa-RT, a model trained on human eye-tracking data to predict reading times at the word level, and validated its performance on both internal and external test sets. We also constructed a dataset of 496 Human–LLM pairs, in which each generated article was obtained by rewriting a human-written original while preserving its informational content as closely as possible. Together, these two contributions allowed us to investigate predicted reading effort at scale across different linguistic realizations of similar content, while reducing the influence of differences in the information conveyed.

Our results show that LLM-generated rewrites tend to be more concise and have slightly lower total reading times than their human-written counterparts, while exhibiting higher average reading times per word. This suggests a trade-off between textual conciseness and processing cost: LLMs convey comparable information using fewer words, but this reduction is accompanied by greater reading effort per word. Moreover, the more extensively LLMs shorten the original text, the larger the increase in reading time per word tends to be. LLM-generated rewrites therefore appear to favor faster reading overall, while concentrating greater processing demands on individual words.

Linguistic characteristics are systematically associated with this pattern. LLM-generated texts tend to use longer and less frequent words, along with greater lexical diversity, all of which are associated with higher reading times per word. At the same time, their words are generally more predictable from the preceding context, a property associated with lower reading times. These findings reveal two opposing tendencies: more demanding lexical choices are associated with greater processing effort, whereas higher contextual predictability appears to partly reduce it. Overall, however, despite their greater contextual predictability, LLM-generated texts still exhibit higher average predicted reading times per word.

Our findings suggest that the cognitive implications of LLM-generated writing cannot be understood solely in terms of textual conciseness. Although LLM-generated rewrites contain fewer words, their linguistic characteristics are associated with higher predicted processing demands per word. This suggests that the efficiency of machine-generated writing should be evaluated not only from the perspective of content production, but also from that of the reader.

Our analysis relies on reading times predicted by a model trained on human eye-tracking data, rather than direct measurements on the Human--LLM pairs. While the results reveal systematic differences in predicted reading demands, further studies involving human participants are needed to establish whether these differences translate into measurable changes in cognitive effort and comprehension.


\section*{Acknowledgment}

This work was partially funded by the European Union within the National Programme 'Research, Innovation and Competitiveness' 2021-2027 - SINTESI Nr. 1\_93098910503\_0000962.


 \bibliographystyle{IEEEtran}
 \bibliography{biblio}

\end{document}